\documentclass{article}
\usepackage{spconf,epsfig}
\usepackage{subfig}
\usepackage{times}
\usepackage{epsfig}
\usepackage{graphicx}
\usepackage{amsmath}
\usepackage{amssymb}
\usepackage{threeparttablex}
\usepackage{xspace}
\usepackage{multirow}
\usepackage{bbding}
\usepackage{multicol}
\usepackage{appendix}
\usepackage{color}
\usepackage{float}
\usepackage{listings}
\usepackage{colortbl}
\usepackage{enumitem}
\usepackage{tablefootnote}
\usepackage[pagebackref=false,breaklinks=true,colorlinks,bookmarks=false]{hyperref}

\usepackage{booktabs}       % professional-quality tables
\usepackage{amsfonts}       % blackboard math symbols
\usepackage{nicefrac}       % compact symbols for 1/2, etc.
\usepackage{microtype}      % microtypography
\usepackage{xcolor}         % colors
\usepackage{natbib}
\setcitestyle{numbers,square}

\newcolumntype{x}[1]{>{\centering\arraybackslash}p{#1pt}}

\newlength\savewidth

\let\OLDthebibliography\thebibliography
\renewcommand\thebibliography[1]{
  \OLDthebibliography{#1}
  \setlength{\parskip}{0pt}
  \setlength{\itemsep}{0pt plus 0.3ex}
}

\begin{document}\sloppy

% Example definitions.
% --------------------
\def\x{{\mathbf x}}
\def\L{{\cal L}}

% Title.
% ------
\title{MRSN: Multi-Relation Support Network for Video Action Detection}
%
% Single address.
% ---------------
\name{Yin-Dong Zheng$^{*}$, Guo Chen$^{*}$, Minglei Yuan, Tong Lu$^{\dag}$
\thanks{$^{*}$These authors contributed equally.}
\thanks{$^{\dag}$Corresponding Author.}
\thanks{This work is supported in part by the National Natural Science Foundation of China (Grant No. 61672273, 61832008).}}
\address{National Key Laboratory for Novel Software Technology, Nanjing University\\
\{zhengyd, cg1177, mlyuan\}@smail.nju.edu.cn\\
lutong@nju.edu.cn}

\maketitle

\begin{abstract}
Action detection is a challenging video understanding task, requiring modeling spatio-temporal and interaction relations.
Current methods usually model actor-actor and actor-context relations separately, ignoring their complementarity and mutual support.
To solve this problem, we propose a novel network called {\em Multi-Relation Support Network} (MRSN).
In MRSN, Actor-Context Relation Encoder (ACRE) and Actor-Actor Relation Encoder (AARE) model the actor-context and actor-actor relation separately.
Then Relation Support Encoder (RSE) computes the supports between the two relations and performs relation-level interactions.
Finally, Relation Consensus Module (RCM) enhances two relations with the long-term relations from the Long-term Relation Bank (LRB) and yields a consensus.
Our experiments demonstrate that modeling relations separately and performing relation-level interactions can achieve and outperformer state-of-the-art results on two challenging video datasets: AVA and UCF101-24.
\end{abstract}
\section{Introduction}
\label{sec:intro}

Spatio-temporal action detection is a challenging task that involves locating the spatial and temporal positions of actors, as well as classifying their actions.
In order to accurately detect and classify actions, spatio-temporal action detection algorithms must take into account not only the human pose of actors but also their interactions with the surrounding context and other actors.
As such, the ability to model and handle actors' interactions in the spatial and temporal dimensions is a key factor of these algorithms.

\begin{figure}[t]
    \centering
    \includegraphics[width=0.9\linewidth]{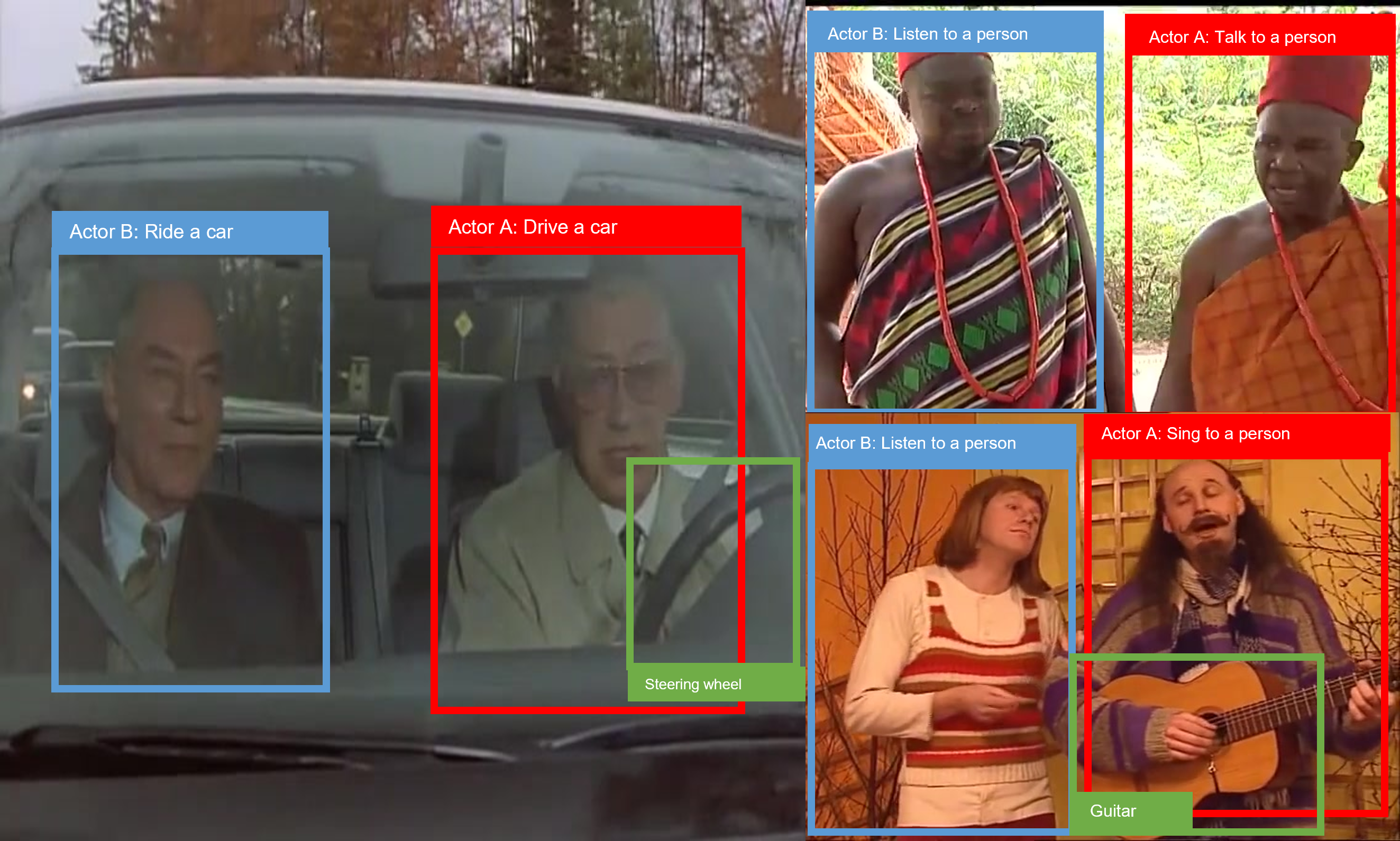}
    \caption{\textbf{Left:} It is difficult to distinguish driver and passenger by separately reasoning the relations between actors and steering wheel. Actor-actor relation supports the distinction between these actions.
\textbf{Right:} In both videos, A opens his mouth while B listens.
Only actor-context relation between the guitar and A reveals A is singing to B.}
    \label{fig: motivation}
\end{figure}

In AVA~\cite{ava}, actions are divided into three broad categories: pose, actor-context action, and actor-actor action.
For actor-context and actor-actor actions, current methods~\cite{ava, lfb, aia, acar} usually model the two relations separately and then fuse them.
We argue that human action semantics are complex and comprise multiple interaction relations.
Separately modeling these two relations ignores complementarity and dependencies between them, making it difficult to extract higher-level interaction semantics effectively.
For example, as shown in Figure~\ref{fig: motivation}, we need actor-actor relations for two actors in a car to distinguish who is driving and who is riding; similarly, distinguishing "sing to a person" and "talk to a person" also relies on actor-context relation.
Therefore, complex relation modeling relies on adaptive complementation and support among multi-relations, which requires not only actor-context and actor-actor interactions but also relation-level interactions.

To tackle the above issues, we propose {\em Multi-Relation Support Network} (MRSN), which focuses on complex relation modeling for action detection.
MRSN takes context and actor features as input.
First, Actor-Context Relation Encoder (ACRE) and Actor-Actor Relation Encoder (AARE) capture the two relations by tokenizing context and actor features and feeding them into the Transformer encoders.
Then, the Relational Support Encoder (RSE) computes the relational support and enriches the details of the two relations.
Finally, Relation Consensus Module (RCM) introduces temporal support information from the Long-term Relation Bank (LRB) and fuses multiple relations to obtain results.

MRSN is validated by extensive experiments conducted on AVA and UCF101-24.
Our contributions are summarized as follows:
1) We propose the concept of relational support and relation-level interaction for actor-context and actor-actor to enrich the details of the two relations.
2) We propose MRSN and introduce LRB to achieve multi-relation and long-short-term relation consensus.
3) MRSN achieves state-of-the-art performance on benchmark datasets.

\section{Related Works}
\textbf{Action Recognition.}
Action recognition aims to classify the actions in the video and can be mainly divided into 2D CNNs and 3D CNNs.
2D CNN~\cite{two-stream, tsn, tdn} uses a 2D convolutional network to extract spatial features while modeling temporal relation by inserting temporal inference modules or extra optical flow.
3D CNN~\cite{i3d,s3d,slowfast} uses 3D convolution to extract features in both temporal and spatial dimensions.
To optimize the backbone, ~\cite{dsn, internvideo} employ training strategies, such as reinforcement learning, as well as generative and discriminative self-supervised learning.
We use SlowFast~\cite{slowfast}, which jointly extracts features through the spatial high-resolution slow and low-resolution fast pathways, as the backbone.

\textbf{Spatio-Temporal Action Localization.}
Spatio-temporal action detection locates actors and classifies their actions.
Some methods~\cite{tube, act} use a single backbone for both tasks.
Most current state-of-the-art methods~\cite{ava, slowfast, acrn, aia, acar, yowo} use a two-backbone pipeline with multi-stage or end-to-end strategy.
This pipeline decouples human detection and action classification, making the model more flexible and suitable for actor-context and actor-actor interaction.
MRSN adopts this two-backbone pipeline.

\textbf{Relational Reasoning and Attention Mechanism.}
Relational reasoning and attention mechanisms play important roles in video understanding.
Non-local~\cite{non-local} insert non-local blocks into the backbone, which compute the response by relating the features at different times or spaces.
LFB~\cite{lfb} applies the non-local operator to long-term feature banks.
~\cite{aia,acar} uses self and cross attention mechanisms to achieve relational reasoning between different instances.
With the development of Transformer~\cite{transformer},~\cite{vivit, timesformer} uses the Transformer's encoder to replace the convolution module in the traditional CNN backbone and extract the relation between pixels.
All modules in MRSN are implemented with Transformer.

\section{Multi-Relation Support Network}
\begin{figure*}[t]
    \centering
    \includegraphics[width=0.9\textwidth]{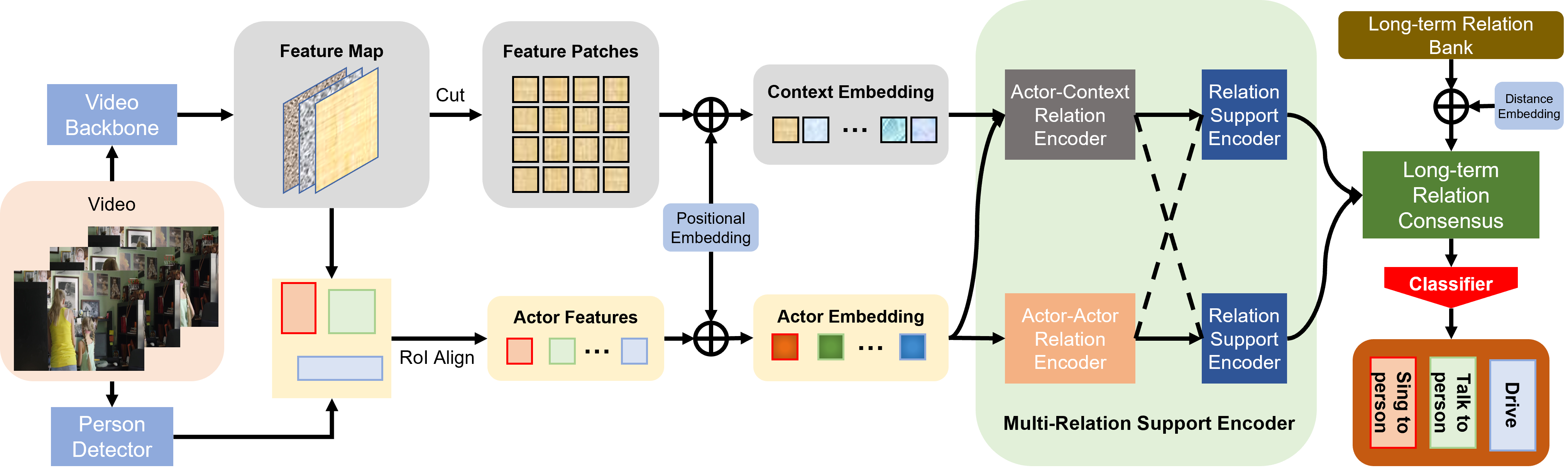}
    \caption{\textbf{Multi-Relation Support Network (MRSN).}
   context and actor-actor relations.
Then, Relation Support Encoder computes relational support between these relations and makes them interact on the relation level.
Finally, the enhanced relations are fused with the long-term relations from the Long-term Relation Bank, yielding a long-short-term relations consensus and classifying.}
    \label{fig:arch}
\end{figure*}

In this section, we describe Multi-Relation Support Network (MRSN) in detail.
As shown in Figure~\ref{fig:arch}, given a clip, an off-the-shelf person detector detects the keyframe and generates candidate boxes, and the backbone extracts a 3D feature map from the clip.
Then, 3D feature map is temporal pooled to 2D feature map $F \in \mathbb{R}^{C \times w \times h}$, and RoI~\cite{fast-rcnn} features ${A^1,A^2,\dots,A^N|A^n\in \mathbb{R}^{C \times 7 \times 7} }$ are extracted for each actor box using RoIAlign~\cite{mask-rcnn}, where $N$ is the number of actor boxes.
After that, MRSM models interaction relations using the feature map and RoI features as input.
Finally, the interaction relations features are fed to an action classifier for multi-label action prediction.

\subsection{Multi-Relation Support Encoder}
The Multi-Relation Support Encoder (MRSE) extracts complex interaction relations.
It models actor-context and actor-actor relations separately but also interacts with them at the relation level to enrich their semantics.
MRSE consists of three sub-encoders: Actor-Context Relation Encoder (ACRE), Actor-Actor Relation Encoder (AARE), and Relation Support Encoder (RSE).
ACRE and AARE model actor-context and actor-actor relations, respectively.
RSE computes relational supports for these relations, which are then fed to subsequent modules for relation-level interaction and semantic complement.
MRSE has a flexible structure, allowing it to be stacked and inserted into other detection heads.

\subsubsection{Actor-Context Relation Encoder}
\label{sec:acre}
Actor-context interactions can be realized in several ways.
ACRN~\cite{acrn} cuts the context feature into a grid, replicates and concatenates the actor feature to all spatial locations, then uses convolution to make the actor interact with context.
Inspired by ACRN's~\cite{acrn} feature gridding and ViT's~\cite {vit} images splitting, we propose Actor-Context Relation Encoder (ACRE), which splits and embeds feature maps into patch sequences to achieve actor-context interactions in the global receptive field.

\textbf{Patch Embedding.}
For the context feature map $F\in \mathbb{R}^{C \times \text{w} \times \text{h}}$, since patch embedding requires feature maps of all clips to have the same shape, we first adaptively pool the feature map to $S \times S$, namely $F'\in \mathbb{R}^{C \times S \times S}$.
Then, we cut $F'$ into $L=\frac{S}{p}\times\frac{S}{p}$ pieces of $p\times p$ patches $\{P_1,\dots,P_L| P_i \in \mathbb{R}^{C \times p \times p}\}$ without overlapping.
The width of the patch $p$ can be any factor of $S$, and in our implementation, we set $S=16, p=2$.
After that, we separately operate Liner Projection on RoI feature $A_n$ and patch sequence $\{P_1,\dots,P_L\}$ to obtain the token sequence $\{R_A,R_1,\dots,R_L|\{R \in \mathbb{R}^{1\times d}\}$, where $R_A$ indicates actor token and $\{R_1,\dots,R_L\}$ indicate the context token sequence.
$d$ is the embedding dimension set to 512 in the experiment.
To retain context positional information and distinguish actor and context, additional learnable actor embedding $E_A$ and positional embeddings $\{E_i|i=1,\dots,L\}$ are added to $R_A$ and $\{R_1,\dots,R_L\}$ respectively.
For each actor in $\{A^1,\dots,A^N\}$, we construct a total of $N$ token sequences $T=\{R_A^n,R_1,\dots,R_L\}_{n=1}^N$.
The entire embedding procedure is shown in Eqn.\ref{eqn:patchembed}.
\begin{equation}
\begin{aligned}
&\{R_1,\dots,R_L\}=\text{PatchEmbedder}(F)\\
&\{R_A^1,\dots,R_A^N\}=\text{ActorEmbedder}(\{A^1,\dots,A^N\})\\
&T^n=\left[R_A^i,R_1,\dots,R_L \right]+\left[E_\text{A},E_1,\dots,E_L \right],\\
&n=1,\dots,N.\\
\end{aligned}
\label{eqn:patchembed}
\end{equation}

\textbf{Actor-Context Relation Encoding.}
For the token sequence $T^n=\{R_A^n,R_1,\dots,R_L\}$ of length $L+1$, we feed it into a Transformer encoder~\cite{transformer} with multi-head self-attention (MSA) to make actor token interact with context tokens.
The encoding procedure can be formalized as follows:
\begin{equation}
\begin{aligned}
X_0&=T^n\\
X_k'&=\text{LN}(\text{MSA}(X_{k-1}))+X_{k-1}\\
X_k&=\text{LN}(\text{FFN}(X_k'))+X_k',\qquad k=1,\dots,M_\text{ACRE}\\
\end{aligned}
\end{equation}
where LN is the LayerNorm layer, and FFN is the Feed Forward Layer using GELU as non-linearity.
$M_\text{ACRE}$ is the number of the encoder.

\subsubsection{Actor-Actor Relation Encoder}
Prior work often pairs RoI features of all actors to extract actor-actor relations or uses self-attention without positional information.
We argue that positional information between actors is beneficial for relation modeling, so we add positional embedding to actor tokens and use the same Transformer Encoder as in Section~\ref{sec:acre} to extract actor-actor relations.

\textbf{Actor Positional Embedding.}
We use the same splitting method in Section~\ref{sec:acre} for the space, that is, the space is split into $L=\frac{S}{p}\times\frac{S}{p}$ patches without overlapping, and we construct $L$ learnable positional embedding $\{E_{i,j},i\in\{1,\dots,\frac{S}{p}\},y\in\{1,\dots,\frac{S}{p}\}\}$.
Then, for each actor box, we calculate its center point, search which patch its center point belongs to, and add the corresponding positional embedding to its token.
Specifically, given an actor box $B^n=\{x_1^n,y_1^n,x_2^n,y_2^n|x,y\in\left[ 0 , 1 \right]\}$, where $(x_1^ n, y_1^n)$ and $(x_2^n, y_2^n)$ are the coordinates of the upper left and lower right corners, respectively.
We assign its positional embedding $E^n_A$ by Eqn.\ref{eqn:actor-pos}.
\begin{equation}
\begin{aligned}
          i^n=\left \lceil \frac{(x_1^n+x_2^n)\times p}{2} \right \rceil &\quad j^n=\left \lceil \frac{(y_1^n+y_2^n)\times p}{2} \right \rceil\\
        E^n_A=&E_{i^n,j^n} \\
\end{aligned}
\label{eqn:actor-pos}
\end{equation}

Finally, the sequence of actor tokens with positional embeddings is fed into the Transformer encoder for self-attention computation of the actor-actor relation.
\begin{equation}
\begin{aligned}
Y_0&=\left[R_A^1,\dots,R_A^N\right]+\left[E_A^1,\dots,E_A^N\right]\\
Y_k'&=\text{LN}(\text{MSA}(Y_{k-1}))+Y_{k-1}\\
Y_k&=\text{LN}(\text{FFN}(Y_k'))+Y_k',\qquad k=1,\dots,M_\text{AARE}\\
\end{aligned}
\end{equation}
where $M_\text{AARE}$ is the number of AARE in each MRSE.

\subsubsection{Relation Support Encoder}
After modeling actor-context and actor-actor relations, RSE will compute supports between the two relations and further make them interact on the relationship level to enrich their semantics.
Inspired by CrossViT~\cite{crossvit}  performs cross-attention on cross-domain information, we use the Bidirectional Cross-attention Transformer to perform relation-level interaction.

Specifically, given the outputs $X$ and $Y$ from ACRE and AARE, we perform bidirectional cross-attention on each actor's $X^n$ and $Y^n$.
The whole procedure is shown in Eqn.\ref{eqn:cross}.
\begin{equation}
    \begin{aligned}
    X_0^n&=X^n,\qquad Y_0^n=Y^n\\
    X_k&=\text{LN}(\text{MCA}_{X}(X_{k-1},Y_{k-1}))+X_{k-1}\\
        Y_k&=\text{LN}(\text{MCA}_{Y}(Y_{k-1},X_{k-1}))+Y_{k-1}\\
    X_k&=\text{LN}(\text{FFN}_{X}(X_{k}'))+X_{k}'\\
        Y_k&=\text{LN}(\text{FFN}_{Y}(Y_{k}'))+Y_{k}' \qquad k=1,\dots,M_\text{RSE}\\
   \end{aligned}
   \label{eqn:cross}
\end{equation}
where $M_\text{RSE}$ is the number of RSE and MCA is the multi-head cross-attention.
The implementation of MCA is shown in Figure~\ref{fig:mca} and Eqn.\ref{eqn:mca}.
\begin{equation}
\begin{aligned}
&\text{MCA}(X,Y)= \text{Concat}(\text{head}_1,\dots,\text{head}_h)W^O\\
&\text{head}_i= \text{Softmax}(\frac{(XW_i^Q)(YW_i^K)^T}{\sqrt{d_k}})YW_i^V\\
\end{aligned}
\label{eqn:mca}
\end{equation}
where $W_i^Q$, $W_i^K$, $W_i^V$, and $W^O$ are the weights of linear transforms, $h$ is the number of heads, and $d_k$ is the dimension of each head.

\subsection{Relation Consensus Module}
The Relation Consensus Module (RCM) fuses actor-context relation $X$ and actor-actor relation $Y$, then yields classification results.
LFB~\cite{lfb} shows long-term information benefits to action classification, so we design two RCMs for short-term (clip) and long-term (clip window) relation consensus.

\subsubsection{Short-Term Relation Consensus}
Short-term RCM ($\text{RCM}_\text{S}$) only yields a consensus of two relations in a single clip.
Given actor-context relation $X^n=\{X_A^n,X_1^n,\dots,X_L^n\}$ and actor-actor relation $Y^n$, where $X_A^n$ is the actor token and $\{X_1^n,\dots,X_L^n\}$ are the context tokens, $\text{RCM}_\text{S}$ performs average pooling on context tokens to obtain context feature $X_C$.
Then, $X_A^n$, $X_C^n$, and $Y^n$ are concatenated and fed to an MLP for action classification.
The short-term relation consensus for actor $n$ can be formulated as Eqn.\ref{eqn:rcms}.
\begin{equation}
    \label{eqn:rcms}
    \begin{aligned}
    X_C^n&=\text{AvgPool}(\left[X_1^n,\dots,X_L^n\right])\\
    G^n&=\text{Concat}(X_A^n,X_C^n,Y^n)\\
    Z^n&=\text{MLP}(G^n)\\
    \end{aligned}
\end{equation}

\begin{figure}[t]
    \centering
    \includegraphics[width=0.8\linewidth]{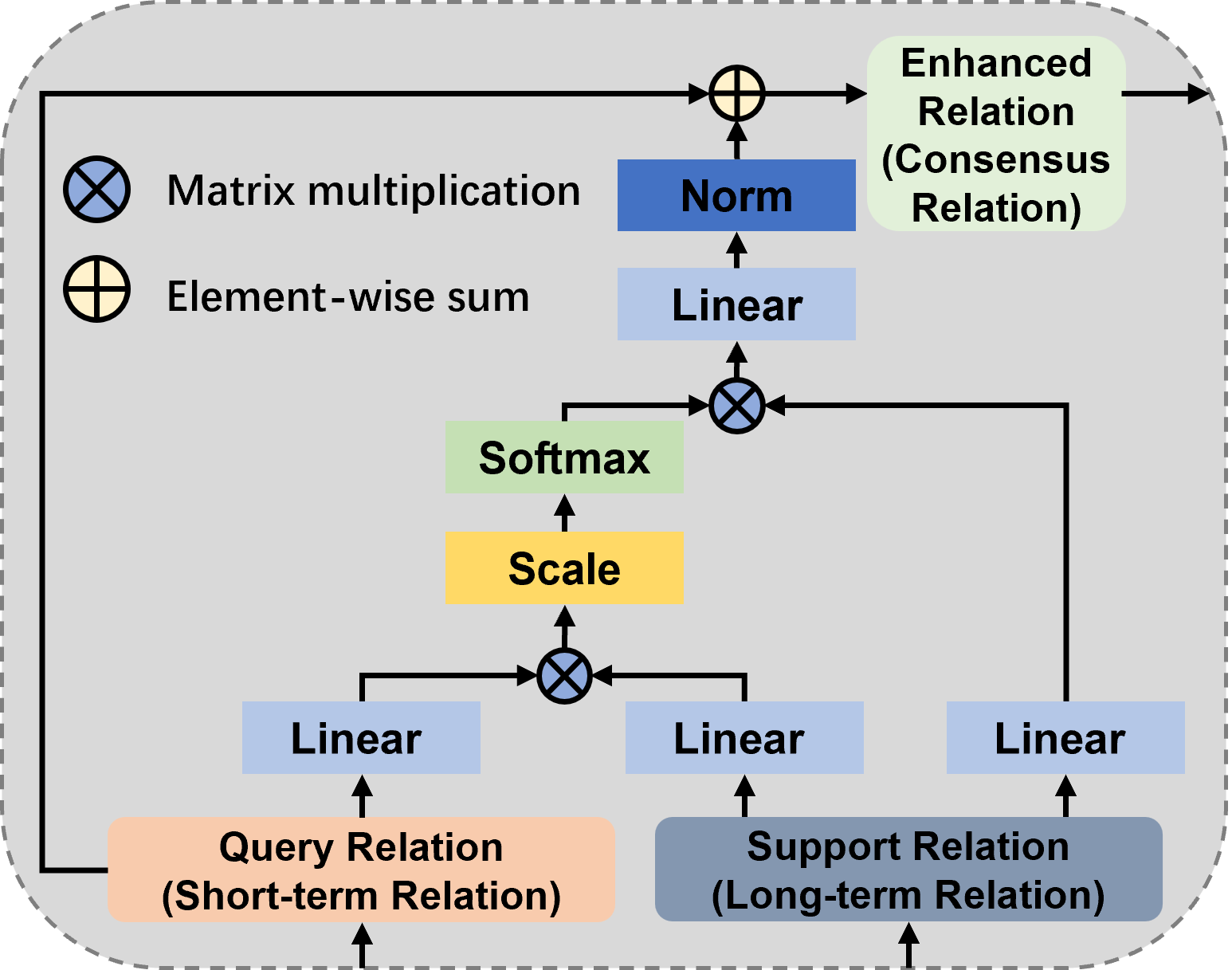}
    \caption{\textbf{Multi-head Cross-attention (MCA) module.}}
    \label{fig:mca}
\end{figure}

\subsubsection{Long-Term Relation Consensus}
To better make short-term information interact with long-term information, we propose Long-term RCM ($\text{RCM}_{L}$) and Long-term Relation Bank (LRB).
First, we train a separate MRSN with $\text{RCM}_{S}$.
Then, given a video, we uniformly sample clips of length 2s with the stride 1s and send these clips to the trained MRSN.
For the output of MRSE, we use the same operation as $\text{RCM}_{S}$ to generate and save feature $\{G_j\}_{j=1}^M$ for each clip, where $M$ is the number of actor boxes in each clip.
We name the $\{G_j\}_{j=1}^M$ of all clips in the video as Long-term Relation Bank, and to avoid confusion, we denote it as $\{\overline{G_j}\}_{j=1}^M$.

After that, we set a temporal window of size $\left[t-\omega,t+\omega\right]$ and take out all clip features in the window from the feature bank to obtain the long-term support features $H_t=\{\{\overline{G_j}\}_{j=1}^M\}_{t-\omega}^{t+\omega}$ for clip $t$.
We give each support feature $\overline{G}$ a learnable distance embedding according to the distance between the clip it belongs to and the clip $t$.
Finally, we take $\{G_n\}_{n=1}^N$ of clip $t$ as Query, $H_t$ as Key and Value, and input them into a Cross-attention Transformer encoder to achieve long-term relation consensus.
The final output is fed into an MLP for action classification.
Since $\text{RCM}_{L}$ is followed by an MLP, we remove the Feed Forward Layer in the Transformer Encoder to save computational overhead.
The Eqn.\ref{eqn:rcml} formalizes this procedure.
\begin{equation}
    Z=\text{MLP}(\text{LN}(\text{MCA}(\{G_n\}_{n=1}^N,H_t+E_\text{dist}))+\{G_n\}_{n=1}^N)
    \label{eqn:rcml}
\end{equation}
where $E_\text{dist}$ is the distance embeddings, and MCA is the same as in Eqn.\ref{eqn:mca} and Figure~\ref{fig:mca}.

\section{Experiments}
\subsection{Datasets}
We evaluate MRSN on AVA and UCF101-24.
AVA~\cite{ava} has 430 videos split into 235, 64, and 131 for training, validating, and testing.
UCF101-24 is a subset of UCF101 with 3,207 videos of 24 sports classes.
Following standard protocol, we evaluate models with frame-mAP by frame-level IoU threshold 0.5.

\subsection{Implementation Details}
\label{sec:detail}
\textbf{Person Detector.}
For person detection on keyframes, we follow previous works~\cite{ava, slowfast} which use the person boxes obtained by the off-the-shelf person detector.
For AVA, the detector is a Faster R-CNN with ResNeXt-101-FPN~\cite{resnext,fpn} backbone, pre-trained on ImageNet~\cite{imagenet} and the COCO~\cite{coco} human keypoint images, then finetuned on the AVA dataset.
For UCF101-24, we use the person detector from~\cite{yowo}.

\textbf{Backbone and MRSM}
We choose the SlowFast~\cite{slowfast} network as the backbone and increase the spatial resolution of res5 by $2\times$.
SlowFast's Slow pathway, named SlowOnly R-50 $16 \times 4$, is used for the ablation study, and SlowFast R-50 \& R-101 for state-of-the-art methods comparison.
For AVA, the input of the backbone is a 64-frame clip, and the slow pathway samples $T = 8$ frames with the temporal step $\tau =8 $, while the fast pathway samples $\alpha T(\alpha=4)=32$ frames.
For UCF101-24, we use SlowFast R-50 $8\times 4$ as the backbone, and the input is a 32 continuous frames clip.
The SlowOnly R-50 and SlowFast R-50 are pre-trained on the Kinetics-400~\cite{kinetics} and SlowFast R-101 is pre-trained on the Kinetics-700.
For ACRE, AARE, RSE, and $\text{RCM}_\text{L}$, the channels are set to 1024, and the number of heads are set to 8.
For $\text{RCM}_\text{L}$, we set $\omega=10$, which means using the relations between 10 seconds before and after the clip, and the window size is 21 seconds.

\begin{table}[]
    \centering
    \caption{Comparison with state-of-the-art methods on AVA v2.1.
\checkmark indicates the method uses optical flow as extra inputs.}
\begin{tabular}{lc|c}
model & flow & mAP \\
\hline
ACRN~\cite{acrn} & \checkmark &17.4\\
SlowFast, R-50~\cite{slowfast} &  & 24.8 \\
SlowFast, R-101~\cite{slowfast} &  & 26.3\\
LFB, R-50~\cite{lfb}  &  & 25.8\\
LFB, R-101~\cite{lfb} &  & 26.8\\
ACAR, R-50~\cite{acar}&&28.3\\
\hline
$\text{MRSN}_\text{S}$, R-50 &  & 27.2 \\
$\text{MRSN}_\text{L}$, R-50 &  & \textbf{28.4} \\
\end{tabular}
    \label{table:sota-ava2.1}
\end{table}

\textbf{Training and Inference.}
We train our models using the SGD optimizer with a batch size of 16, a weight decay of $10^{-7}$, a momentum of 0.9, and the linear warm-up strategy.
For AVA, the models are trained for 6 epochs with a learning rate of 0.04, which is decreased by 10 at epochs 5.6 and 5.8.
During training and inference, we scale the shorter side to 256.
In training, we use ground-truth boxes and proposals with an IoU greater than 0.75 as training samples.
In inference, person boxes with scores greater than 0.85 are used for action classification.
For UCF101-24, the models are trained for 5 epochs, with the learning rate decreased at epochs 4.8 and 4.9.

\begin{table}[]
    \centering
    \caption{Comparison with state-of-the-art methods on AVA v2.2.}
\begin{tabular}{lc|c}
model & pretrain & mAP \\
\hline
SlowOnly, R-50~\cite{slowfast}&  K400 & 20.9\\
SlowFast, R-50~\cite{slowfast}&  K400 & 25.6\\
SlowFast, R-101~\cite{slowfast} & K600 & 29.1\\
AIA, R-50~\cite{aia}& K700 & 29.8\\
AIA, R-101~\cite{aia}& K700 & 32.3\\
ACAR, R-101~\cite{acar} & K700 & 33.3\\
\hline
$\text{MRSN}_\text{S}$, R-50 &  K400 & 28.3 \\
$\text{MRSN}_\text{L}$, R-50 &  K400 & 29.2 \\
$\text{MRSN}_\text{S}$, R-101 & K700 & 32.1\\
$\text{MRSN}_\text{L}$, R-101 & K700 & \textbf{33.5}\\
\end{tabular}
    \label{table:sota-ava2.2}
\end{table}

\subsection{Comparison with the State-of-the-art Methods}
We compare MRSN with state-of-the-art methods on the validation set of AVA v2.1 (Table~\ref{table:sota-ava2.1}) \& v2.2 (Table~\ref{table:sota-ava2.2}) and the UCF101-24 split-1 (Table~\ref{table:sota-ucf}).
LFB, AIA, ACAR, and MRSN all use SlowFast as the backbone. 
$\text{MRSN}_\text{S}$ and $\text{MRSN}_\text{L}$ indicate MRSN equipped with $\text{RCM}_\text{S}$ and $\text{RCM}_\text{L}$.
Since the videos in UCF101-24 are very short, we only report the result for $\text{MRSN}_\text{S}$.
For a fair comparison, we do not use multi-model ensemble or multi-scale augmentation.
All results are based on single models and scale 256.

\textbf{AVA.}
On AVA v2.1, compared to baseline SlowFast, R-50, $\text{MRSN}_\text{S}$, R-50 improves the mAP by 2.4.
$\text{MRSN}_\text{L}$ with Long-term Relation Bank outperforms all state-of-the-art models.
Although $\text{MRSN}_\text{L}$ is only 0.1 higher than ACAR, the amount of parameters and computation of MRSN is only half of that in ACAR.
On AVA v2.2, $\text{MRSN}_\text{L}$ with backbone SlowFast R-50 pre-trained on Kinetics-400 is higher than SlowFast R-101 pre-trained on Kinetics-600.
$\text{MRSN}_\text{L}$, R-101 is 1.2 higher than AIA, R-101.
It is worth noting that AIA uses an object detector for object detection, which means an extra instance detector and more computational overhead.

\begin{table}[]
    \centering
    \caption{Comparison with state-of-the-art methods on UCF101-24 split1.
\checkmark indicates extra input optical flow.}
\begin{tabular}{lc|c}
model & flow & mAP \\
\hline
ACT~\cite{act} &  & 69.5\\
STEP, I3D~\cite{step} & \checkmark & 75.0\\
C2D~\cite{aia} &  & 75.5\\
Gu {\em et al.}~\cite{ava} & \checkmark & 76.3\\
AIA, R-50~\cite{aia}&  & 78.7\\
S3D-G~\cite{s3d} & \checkmark & 78.8\\
\hline
$\text{MRSN}_\text{S}$, R-50 &  & \textbf{80.3}\\
\end{tabular}
    \label{table:sota-ucf}
\end{table}

\textbf{UCF101-24.}
On UCF101-24, MRSN outperforms all methods (including two-stream methods).
Since UCF101-24 has less than 1.4 actors per video on average, action classification does not rely on actor-actor interaction, mainly evaluating actor-context modeling ability.
MRSN's mAP is 1.6 higher than AIA using an additional object detector, demonstrating its strong ability to capture actor-context interactions.

\begin{table}[t]
    \centering
        \caption{Ablation on eash component of MRSN.}
\begin{tabular}{cccc|c}
\toprule
ACRE&AARE&RSE& LRB & mAP \\
\hline
 & &&&20.94 \\
 \checkmark&&&& 22.35\\
&\checkmark&&& 21.69\\
\checkmark&\checkmark&&& 22.89\\
\checkmark&\checkmark&\checkmark&& 23.21\\
\checkmark&\checkmark&\checkmark&\checkmark& \textbf{24.22}\\
\bottomrule
\end{tabular}

    \label{table:comp}
\end{table}

\subsection{Ablation Study}
We conduct ablation studies on AVA v2.2 using SlowOnly as the backbone. 
We set the channel dimension of all encoders to 1024 and the number of heads in the Transformer to 8.
We only do component analysis in the paper and put the rest of the ablation studies in the supplementary material.

To validate the effectiveness of the MRSN design, we ablated each component of the MRSN.
As shown in Table~\ref{table:comp}, the performance of AARE is worse than ACRE because AARE only takes RoI features as input, and the lack of context information will seriously impair the classification of actor-context actions.
Although ACRE only performs actor-context interaction, the context features contain actors' features, which helps classify actor-actor actions.
Compared with single relation modeling, the fusion of the two relations has achieved significant performance improvements.
RSE brings a performance improvement of 0.32, demonstrating the effectiveness and importance of relation-level interaction.
LRB can additionally improve mAP by 1.01,  confirming that the integration of long-term information is beneficial to action detection.

\section{Conclusion}
In this paper, we analyzed the complexity of action relation semantics and proposed Multi-Relation Support Network (MRSN).
MRSN enhances and fuses actor-context and actor-actor relations by modeling relations separately and performing relation-level interaction.
We also proposed a Long-term Relation Bank (LRB) to integrate the long-term video information.
Experiments and ablation studies validated the effectiveness and efficiency of MRSN design.
However, relational support and relation-level interaction are currently performed at clip level instead of video level and are not used for human detection refinement.
Applying these techniques to video-level action tubes remains an open and challenging problem.

% -------------------------------------------------------------------------
\bibliographystyle{IEEEbib}
\bibliography{icme2023template}

\end{document}